# Detecting automatically the layout of clinical documents to enhance the performances of downstream natural language processing

Authors and affiliations:

**Christel Gérardin\*, MD :** Sorbonne Université, Inserm, Institut Pierre-Louis d'Epidémiologie et de Santé Publique, Paris, France F7501, Innovation and Data unit, IT Department, Assistance Publique-Hôpitaux de Paris, Paris, France
**Perceval Wajsbürt\*, PhD :** Innovation and Data unit, IT Department, Assistance Publique-Hôpitaux de Paris, Paris, France
**Basile Dura :** Innovation and Data unit, IT Department, Assistance Publique-Hôpitaux de Paris, Paris, France
**Alice Calliger :** Innovation and Data unit, IT Department, Assistance Publique-Hôpitaux de Paris, Paris, France
**Alexandre Mouchet :** Innovation and Data unit, IT Department, Assistance Publique-Hôpitaux de Paris, Paris, France
**Xavier Tannier**, PhD : LIMICS, Sorbonne Université
**Romain Bey, PhD :** Innovation and Data unit, IT Department, Assistance Publique-Hôpitaux de Paris, Paris, France
**\*both authors contributed equally**

# Abstract

**Background:**
The use of clinically derived data in secondary use within health data warehouses for research or steering purposes can be complex, especially when analyzing textual documents from PDFs provided by the source software.

**Objective:**
Develop and validate an algorithm for analyzing the layout of PDF clinical documents to improve the performance of downstream natural language processing tasks.

**Materials and Methods:**
We designed an algorithm to process clinical PDF documents and extract only clinically relevant text. The algorithm consists of several steps: initial text extraction using a PDF parser, followed by classification into categories such as body text, left notes, and footers using a Transformer deep neural network architecture, and finally an aggregation step to compile the lines of a given label in the text. We evaluated the technical performance of the body text extraction algorithm by applying it to a random sample of documents that were annotated. Medical performance was evaluated by examining the extraction of medical concepts of interest from the text in their respective sections. Finally, we tested an end-to-end system on a medical use case of automatic detection of acute infection described in the hospital report.

**Results:**
Our algorithm achieved per-line precision, recall, and F1 score of 98.4, 97.0, and 97.7, respectively, for body line extraction. The precision, recall, and F1 score per document for the acute infection detection algorithm were 82.54 (95CI 72.86-91.60), 85.24 (95CI 76.61-93.70), 83.87 (95CI 76, 92-90.08) and 80.35 (95CI 70.08-89.75), 73.77 (95CI 62.71-83.61), 76.92 (95CI 69.09-84.8), with or without exploitation of the results of the advanced body extraction algorithm, respectively.

**Conclusion:**
We have developed and validated a system for extracting body text from clinical documents in PDF format by identifying their layout. We were able to demonstrate that this preprocessing allowed us to obtain better performances for a common downstream task, i.e., the extraction of medical concepts in their respective sections, thus proving the interest of this method on a clinical use case.

# I. Introduction

## Background

In recent years, electronic health records (EHRs) stored in large clinical data warehouses have become widely available. Health databases, such as those of the Assistance Publique - Hôpitaux de Paris (AP-HP), have facilitated the secondary use of clinical notes for epidemiological research, pharmacovigilance, automatic detection of patient cohorts and the development of diagnostic or therapeutic prediction models. One of the challenges of these databases is to process a very large volume of documents: currently more than 120 million at the AP-HP. The automated analysis of these clinical notes has been made possible by natural language processing (NLP) algorithms, which are particularly adept at extracting named entities of interest - such as medications, symptoms, comorbidities and diagnostic procedures -, text classification, translation,etc.
However, NLP algorithms are often designed to be applied on plain text, but in many health databases, due to an imperfect interoperability of many clinical softwares, documents are primarily available only as PDFs whose layout depend on the clinical software from which they originate. Prior to the extraction of named entities of interest- or other NLP tasks, the plain text information is often derived from the direct capture of PDF documents using a simple mask, introducing noise and decreasing the performance of textual information extraction. This process often leads to the loss of the document structure, in particular regarding section layout. To address this issue, which is not specific to medical documents, several teams have proposed methods for joint analysis of document layout and corresponding text for enhanced comprehension.
Liu et al. [Liu2019] presented a graph convolution method for multimodal information extraction in visually rich documents, utilizing a combination of graph embeddings for layout encoding and text embeddings with a BiLSTM-CRF, outperforming models based solely on textual information. Xu et al [Xu2020, Xu2022] and Huang et al. [Huang2022] introduced three successive methods - LayoutLM, LayoutLMv2, and LayoutLMv3 - for automatic document analysis that integrate both text and layout information into a single model. These templates have achieved state-of-the-art results, with LayoutLMv3 outperforming the others. LayoutLM is based on the BERT architecture, incorporating 2D layouts and embedded images, while the LayoutLMv2 and LayoutLMv3 versions use a multimodal Transformer architecture that incorporates text, layout and image information.
For all models, the initial extraction of text and layout is conducted using optical character recognition (OCR) or a PDF parser. Other methods have also been proposed [Majmuder2020, Kim2022, Yu2021].

One of the challenges of these information extractions is to respect the architecture of the text and therefore the titles of the sections, restored in the right order (usually corresponding to "reason of admission", "medical history", "usual treatment", etc…). Several EHR text analysis can benefit from section identification: enabling a temporal relation extraction [Kropf2017], abbreviation resolution [Zweigenbaum2013], cohort retrieval [Edinger2017]. Automatic or semi-automatic section identification in narrative clinical notes has been studied in the past, as shown by Pomares et al. in a recent review paper [Pomares2019]. They define section identification as detecting the boundary of sections in the text. A section generally corresponds to a paragraph summarizing one dimension of the patient (medical history, allergies, physical exam, evolution during its hospital stay, usual treatments, discharge treatments, *etc.*). According to this review, the majority of the studied papers (59% for 39 studies analyzed) used rule-based methods, 22% machine learning methods and 19% both. Authors also highlight that very few studies presented results with a formal framework. Finally almost all the studies relied on a custom dictionary.

# Goal of the study

In this study, we propose an end-to-end algorithm that processes standard clinical documents in PDF format by extracting the body text separately from the left-hand notes, footnotes, signatures, and other elements, retaining only the clinically relevant content and preserving its original structure, thus correctly detecting the sections. The objective of the algorithm is to be applied to a very large number of documents (more than 120 million) and therefore must be very lightweight.
We demonstrate that this main body text detection step significantly enhances the performance of downstream information extraction tasks. As a proof of concept, we evaluate the performance of an algorithm designed to automatically identify acute infections in clinical documents, utilizing a named entity recognition algorithm and a section identification based on a rule-based entity classification.

# II. Material and Methods
## A. Datasets

For this study we had two datasets at our disposal described below:
- a first dataset of 272 annotated PDF documents for the development and technical validation of the text-extracting algorithm.
- a cohort of auto-immune diseases patients (with systemic lupus erythematosus, Takayasu disease, scleroderma and antiphospholipid syndrom) with 151 fully annotated document with medical concepts and 200 with or without a phenotype of interest, respectively for the medical validation and the illustrative use case.

This study was approved by the local institutional review board (IRB00011591, decision CSE22-18). Subjects that objected to the reuse of their data were excluded. For privacy protection reasons, researchers do not have direct access to the PDF documents. A text-extracting algorithm and a text-pseudonymization algorithm were consequently applied to all the documents before any delivery [Tannier2023].

1. Hospitals' clinical documents (development and technical validation)

The wide majority of clinical reports, including discharge summaries, imaging reports, emergency department reports, surgical reports, prescriptions, pathology reports, and more, are imported into the AP-HP clinical data warehouse as PDF files. A total of 272 randomly selected reports were sampled, ensuring a 3:1 weighting in favor of the following note types: consultation reports, hospitalization reports, operative reports, pathology reports, imaging reports, discharge letters, procedure reports, prescriptions, and emergency department visit reports. These reports were annotated by three independent annotators who segmented the PDFs by marking boxes of interest. These boxes encompassed body text, footers, headers, left notes, page indices, signatures, titles, and a "others" category for elements that did not fit the previous categories. The annotated bounding boxes were then aligned with the lines from the PyMuPdf v1.21.0 parser [PyMuPDF], as illustrated in Figure 1, to create a supervised line classification corpus. This dataset was subsequently divided into a training set consisting of 215 documents and a test set consisting of 57 documents, the number of corresponding line annotations can be found in Supplementary Material, Table 1.

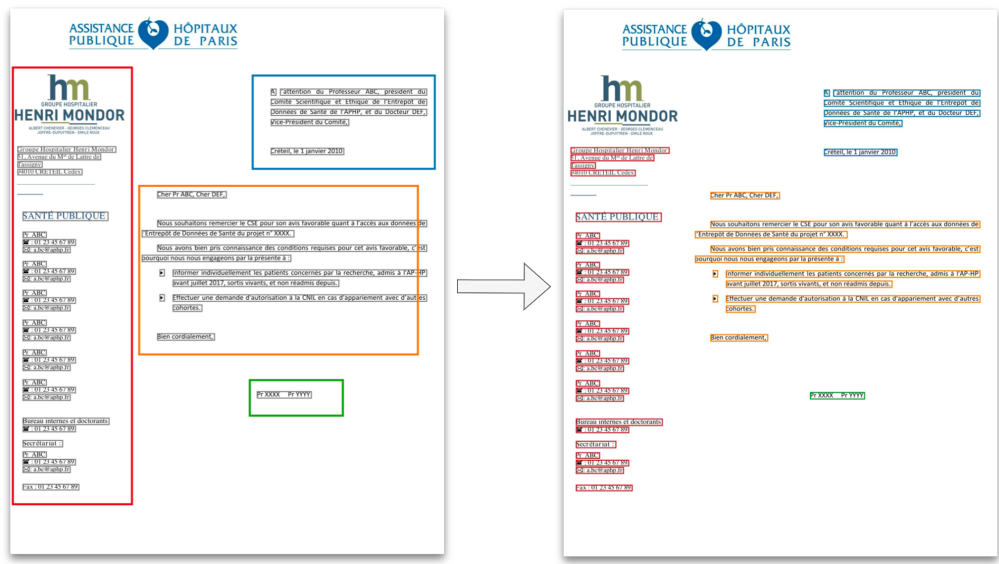

**Figure 1**: Alignment process between annotated bounding boxes and extracted lines using the PyMuPdf parser. The figure illustrates how the different categories of bounding boxes are aligned, enabling the creation of a supervised line classification corpus for further training and analysis.

2. Auto-immune diseases cohort's discharge summaries (medical validation and illustrative use case)

The auto-immune diseases cohort also comes from the AP-HP data warehouse and is restricted to four diseases: Systemic lupus erythematosus, Takayasu disease, scleroderma and antiphospholipid syndrome. This cohort was chosen because of its availability and the fact that an important dataset was already manually annotated.

We had two distinct datasets from this cohort:
- A first set of 151 hospitalization reports, with annotated medical concepts of four different UMLS semantic types: "*Chemicals and drugs*" (e.g., acetaminophen), "*Disorders*" (e.g., meningioma), "*Signs and symptoms*" (e.g., fever, headache), and "*Procedures*" (e.g., brain MRI), and also with vitals parameters present in the text (e.g. body temperature, blood pressure etc.). These reports come from We chose this dataset giving its availability and given the diversity of signs and symptoms. Within this document set, sections ("medical history," "medications," "conclusion," etc...) were also annotated.
This dataset was used to train (80%) and test (20%) the named entity recognition algorithm and section identification. The documents were annotated by a physician (C.G.), after a naive body text extraction described below.

- A second set of 200 hospital reports of other patients randomly selected in the auto-immune cohort dataset was annotated by a physician (C.G.) to indicate the presence or absence of an acute infection in the text. The definition of an "acute" infection was considered broadly: it could be bacterial, fungal, viral, or parasitic and could be an active or uncontrolled chronic infection (e.g., an acute complication of HIV disease) or a new infection (e.g., a pulmonary infection requiring hospitalization). In keeping with the usual clinical convention, an infection mentioned in the "clinical progress", "conclusion" or "reason for hospitalization" sections of the report was labeled as an acute infection, while those mentioned in the "medical history" section were considered as old, chronic controlled infections. (e.g., "had pertussis in childhood").

## B. Algorithms' architectures

The vast majority of clinical documents contained in the Health data Warehouse at the AP-HP are exports from one of the various software programs used by clinicians, and scanned documents make up a negligible fraction (less than 10%) of the total documents. Therefore, we do not resort to Optical Character Recognition (OCR) from rendered documents and focus on text-based document instead.

### Naive algorithm

Our baseline algorithm, which is the one previously used at the AP-HP CDW (Clinical Data Warehouse), generates plain texts by applying a simple mask to the patient documents in PDF format and a pseudonymisation model. This method, while requiring minimal training data (just enough to manually design the mask) and computational resources, effectively extracts the text body from simple documents that conform to the layout used when designing the mask. This method may work in the most common cases, but it does not accommodate different PDF layouts and sometimes produces a mix of administrative information (e.g. dates, hospital wards) and clinically relevant information. In addition, this mask sometimes results in the loss of the original structure of the document, with elements such as section headings being merged at the top of the extracted text. Figure 2 shows an example of the previous text extraction method.

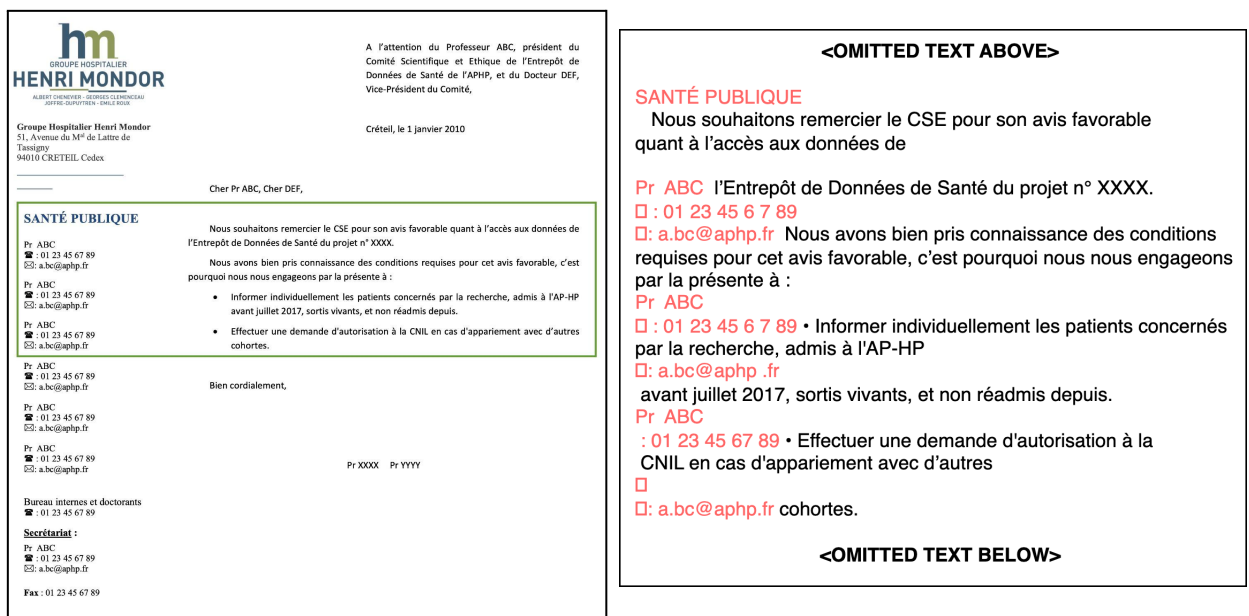

**Figure 2**: Illustration of a failure-case of the naive extraction algorithm. The text snippet shown on the right comes from the green rectangle superimposed on the PDF document on the left. The rules did not correctly identify the side note, resulting in a confusing blend of pertinent text and administrative details, marked in red.

## Advanced algorithm

The end-to-end system comprises a PDF parser to extract lines of text in a PDF, a classifier to infer the type of each line, and an aggregator to compile lines of given labels together to obtain the final textual output. The overall system is illustrated in Figure 3. We chose the PyMuPDF v1.21.0 library [PyMuPDF] to perform the line extraction from the PDF and PyTorch [PyTorch] v1.12.1 to implement the neural network. The aggregation is performed by sorting lines of a given label in a top-down left-right fashion and concatenating them.

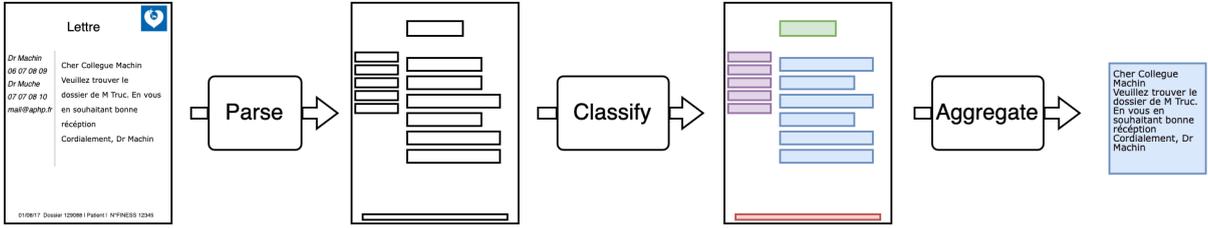

**Figure 3**: Overall system architecture.

## Deep-learning model

The architecture consists of a non-pretrained Transformer that performs classification on extracted lines from the PDF. Each line is represented by a sum of a textual and a layout embedding encoded with 96 dimensions. The textual embedding is a pooled convolution window of 3, 4 and 5 tokens over some embedded text features of the tokens in each line: their 3-letter prefix, 3-letter suffix, shape, and normalized variant, as illustrated by Figure 4. The layout embedding is a concatenation of sinusoidal encodings, introduced by Vaswani (2017) [Vaswani2017], of the left horizontal position, right horizontal position, top vertical bound, bottom vertical bound, width, and height.
Each line representation is then contextualized through a 4-layer Transformer with self-attention. Additionally, we inject the relative 2D distances between lines inside the attention mechanism using a similar mechanism to He et al. (2020) [He2020]. This attention is the sum of content-content attention (the standard dot product attention), content-position attention, and position-content attention:

$$\begin{aligned}
\text{attention}(u,v) = & \frac{(W_1^c u) \cdot (W_2^c v)^T}{\sqrt{3d}} & \text{content to content} \\
& + \frac{([W_1^{dx} dx_{u \to v}; W_1^{dy} dy_{u \to v}]) \cdot (W_2^c v)^T}{\sqrt{3d}} & \text{content to position} \\
& + \frac{(W_1^c u) \cdot ([W_2^{dx} dx_{u \to v}; W_2^{dy} dy_{u \to v}])^T}{\sqrt{3d}} & \text{position to content}
\end{aligned}$$

with $W_1^c, W_2^c, W_1^{dx}, W_1^{dy}, W_2^{dx}, W_2^{dy}$ six projection matrices, $dx_{u \to v}$ the embedding of the relative horizontal position of $v$ w.r.t. $u$ and $dy_{u \to v}$ the embedding of the relative vertical position of $v$ w.r.t. $u$.

Finally, each contextualized line embedding is forwarded through a linear layer followed by a softmax to compute the probabilities of each line's label. The complete architecture is described in Figure 4.

The model, containing 3 million parameters, was trained for 1,000 steps by minimizing the cross-entropy loss between the line's label logits and their gold-annotations using the Adam optimizer. The learning rate was scheduled with a warmup of 100 steps followed by a linear decay from 1e-4 to 0. The model architecture and hyperparameters were selected manually by maximizing the F1-score on a 90-10% train-dev split on the corpus full training set.

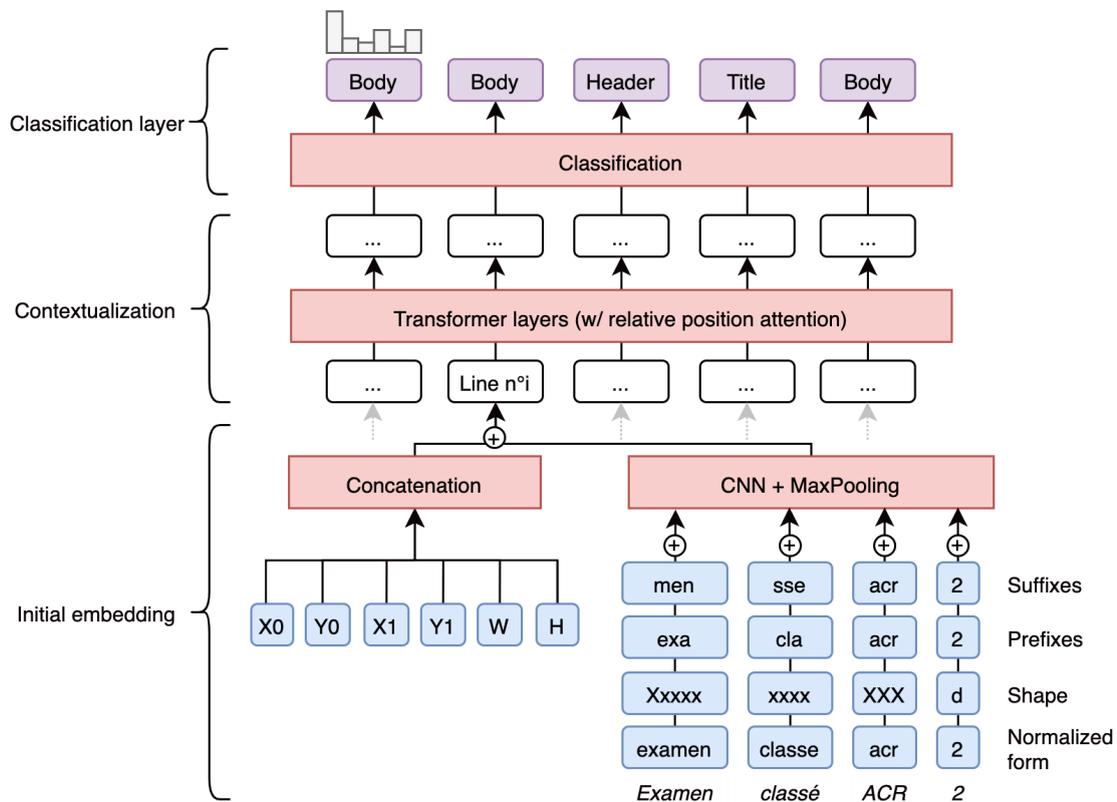

**Figure 4**: Architecture of the deep-learning line classification model. Textual and layout features of each line are embedded to obtain a single representation per line, and are then contextualized with a 4-layer Transformer using self-attention with relative position information. Lastly, they are classified using a linear layer and a softmax function to obtain the probability of each label.

Our implementation is available on github: https://github.com/aphp/edspdf (v0.5.3).

## C. Algorithms' development and technical validation

To validate the performance of our PDF extraction algorithm, we conducted an evaluation using various performance metrics, including precision, recall, F1-score, with micro- and macro-averages. We assessed the performance of the algorithm on the test set of the annotated PDF dataset against the different labels, i.e. body, footer, header, left note, page, others, signature, and title.

Additionally, we performed ablation studies to understand the contribution of different components in the model. These studies involved the removal of the relative position information from the attention and the removal of the whole Transformer layer, and their effects were observed on micro-and macro-averages of the F1-score, body F1-score, and body recall (resp. F1-score and recall for the *body* section identification). We report the results as the mean over 5 runs with different weight initialization seeds.

### D. Medical validation : medical concepts-detection in their respective sections

Key information in clinical documents is conveyed by medical concepts directly present in the text. These medical concepts are present in their respective sections allowing to explain the clinical reasoning (entry treatments, discharge treatments, personal and family comorbidities, etc.). The hypothesis of our study is that a better extraction of the body text of the PDF - which separates the administrative and non-clinical information - can lead on the one hand to a better extraction performance of the clinical concepts directly and on the other hand to a better extraction of the sections to which these concepts belong.

#### 1. Medical concept extraction and classification

Medical concepts can be classified on semantic types as follows as proposed by [McCray2001]: *Chemicals, Anatomical structures, Concept and ideas, etc*. These concepts retain the key information we want to extract and have been annotated with four semantic types : *Chemicals and drugs, Signs and symptoms, Diseases and Procedure* on the Autoimmune Disease Cohort dataset, as shown in Figure 5.

[Annotated medical document excerpt — Autoimmune Disease Cohort]

: réintroduction du METHOTREXATE SC pour réapparition des synovites.

Pelade décalvante traitée par 5 bolus de corticothérapie de ▮ à ▮ et méthotrexate depuis ▮

Docteur ▮

2/7
CRH service SAT CARDIOLOGIE, Imprimé le ▮ Pat.: ▮

EXAMEN CLINIQUE A L'ENTRÉE
TRAITEMENT A L'ENTRÉE
RESUME CLINIQUE - HISTOIRE DE LA MALADIE

MESALAZINE 1g 3 comprimés matin au cours des repas à partir du ▮
COLECALCIFEROL 100000 UI/2mL
CLOBETASOL
FOLIQUE ACIDE 3 comprimés matin chaque samedi

**Figure 5**: Example from the Autoimmune Disease Cohort, annotated with named entities that fall into one of the following four semantic types: *Chemicals and drugs, Signs and symptoms, Diseases and Procedure.*

The automatic extraction of these concepts – also called entity recognition – is performed by our previously described algorithm [Wajsbürt2021, Gérardin2022]. It is based on an encoder and decoder architecture, similar to that of [Yu2020]. Word representations are computed by concatenating output embeddings from the FastText model [Bojanowski 2017], a BERT Transformer [Devlin2018], and a character-based CNN (Convolutional Neural Network). These word representations are then re-contextualized using a multi-layer Bi-LSTM. Ultimately, named entities are identified by applying multiple conditional random fields (CRF, one for each label) on the text and disambiguating overlapping entities with the same label by matching the beginning and end boundaries using a biaffine matcher.
This algorithm and its performance at entity level is assessed on the 151 annotated documents with and without the advanced layout detection algorithm. The metrics used for performance evaluation are precision, recall and F1 score.

Finally, we used our multilabel medical concept classifier to classify all symptoms and disorders in the main medical domains (cardiology, neurology, etc..) -corresponding to the MeSH[MeSH]-C headings- [Gérardin2022]. Specifically in our "Acute infection" use case, this algorithm predicts whether a disease is an infection or not.

2. Section identification

Like previous authors [Pomares2019], we assumed that all sections are preceded by section titles (which was the case in 150 over 151 documents). For the detection of the section title, a custom dictionary was created. The level of granularity of sections versus subsections was discussed collegially and, drawing on previous work [Pomares2019], a set of 14 sections of interest was selected (see Table 1 and Supplementary data Table 2 for the synonyms dictionary).

| | |
|---|---|
| *Antécédents* | History |
| *Antécédents familiaux* | Family history |
| *Allergies* | Allergies |
| *Mode de vie* | Lifestyle |
| *Traitement* | Treatment |
| *Traitement entrée* | Treatment at admission |
| *Traitement sortie* | Discharge treatment |
| *Motif* | Motive (Reasons of admission) |
| *Histoire de la maladie* | History of the actual disease |
| *Evolution* | Clinical Progress |
| *Examen clinique* | Physical examination |
| *Constantes* | Vitals |
| *Examens complémentaires* | Complementary investigations |
| *Conclusion* | Conclusion |

**Table 1**: List of section types to be identified in a clinical report.

For simplicity and clarity, section identification was based on rules, taken from the custom dictionary, directly searching for exact mentions of the section title or synonyms. The method was tested on the auto-immune diseases cohort dataset with and without the advanced layout detection algorithm (described in Section II.A). The section types were all manually annotated by a clinician (C.G.) in the 151 documents extracted by the naive mask. When the structure was not explicit, the section starts were inferred directly from the clinical information (e.g., family history for "Diabetes in brother and sister", etc.). We evaluate the ability of the system to extract the correct medical concepts in the correct sections of interest. The metrics used to assess performance were precision, recall and F1 score.

### E. Illustrative use case: automatic detection of acute infections in the text.

Finally, we wanted to illustrate how our advanced body extraction algorithm, combined with section extraction and medical concept recognition and classification, could automatically detect acute infections in medical reports. Acute infection was chosen since it is a frequent disorder in auto-immune patients treated with corticosteroids or other immunosuppressive drugs. The global approach described above was used: first, the body text of the report was extracted, then, the sections were identified, and finally the concepts in the "Evolution" and "Conclusion" sections were extracted and classified. A concept classified as *infection* in the "Evolution" or "Conclusion" sections categorizes the patient as having an acute infection (including chronic decompensated infections requiring acute management).

# III. Results

## A. Text-extraction and line classification results (technical validation)

Table 2 displays the per-line micro-averaged precision, recall, and F1-score for our text-extraction and classification algorithm. The algorithm achieved a precision of 0.98, a recall of 0.97, and an F1-score of 0.98 for the "body" lines, and an overall micro-average of 0.96 for lines of all types.
Table 3 presents the findings of the ablation study, illustrating a decline in performance for both ablations. The performance degradation of the simplest, non-contextualized model ranges from approximately 2 points in body recall, and body F1-score, 4.5-point decrease in micro-averaged F1-score, to an almost 6-point drop in macro-averaged score.

| Label | Precision | Recall | F1-Score |
|---|---|---|---|
| body | 0.98 | 0.97 | 0.98 |
| footer | 0.84 | 0.88 | 0.85 |
| header | 0.90 | 0.95 | 0.93 |
| left_note | 0.98 | 0.99 | 0.99 |
| page | 0.96 | 0.91 | 0.94 |
| others | 0.97 | 0.92 | 0.94 |
| signature | 0.87 | 0.80 | 0.83 |
| title | 0.87 | 0.80 | 0.84 |
| ALL (macro-avg) | 0.92 | 0.90 | 0.91 |
| ALL (micro-avg) | | 0.96 | |

**Table 2**: Per-line precision, recall and F1-score of the body-extraction algorithm for the test set and for each line type (technical validation).

| Model | Micro-avg F1 | Macro-avg F1 | Body F1 | Body recall |
|---|---|---|---|---|
| Full model | 0.96 | 0.91 | 0.98 | 0.97 |
| -Relative position attention | 0.95 | 0.89 | 0.97 | 0.96 |
| -Transformer | 0.92 | 0.85 | 0.95 | 0.95 |

**Table 3**: Ablation study of model architecture.

## B. Section and medical concepts extraction results (medical validation)

Results on entity extraction in their respective sections are shown in Table 4. Only pairs (entities/sections) of medical interest were kept (i.e. Drugs in the sections "drugs at entry" and "drugs at discharge", Symptoms in the section "physical examination" etc..) . The entities are annotated and the sections are extracted with a rule-based algorithm (and compared to a gold standard annotation). We compared both extractions: with the naive body extraction algorithm, a baseline mask, and with the advanced body extraction and noticed an overall improvement of 0.1 for the F1 score of the entities detection within their right respective sections.

|  | Naive body extraction | | | Advanced body extraction | | |
| --- | --- | --- | --- | --- | --- | --- |
| **Entity**<br>*Corresponding section* | **Precision** | **Recall** | **F1** | **Precision** | **Recall** | **F1** |
| Drugs<br>*at entry* | 0.68 | 0.99 | 0.81 | 0.91 | 0.98 | **0.94** |
| Drugs<br>*at discharge* | 0.82 | 0.91 | 0.86 | 0.89 | 0.99 | **0.94** |
| Vital parameters<br>*physical examination* | 0.84 | 0.93 | 0.88 | 0.94 | 0.99 | **0.96** |
| Procedure<br>*investigations* | 0.83 | 0.98 | 0.90 | 0.87 | 0.98 | **0.92** |
| Signs and Symptoms<br>*medical history* | 0.79 | 0.65 | 0.71 | 0.90 | 0.81 | **0.86** |
| Signs and Symptoms<br>*family medical history* | 0.71 | 0.99 | 0.83 | 0.80 | 0.98 | **0.88** |
| Signs and Symptoms<br>*conclusion* | 0.73 | 0.52 | 0.61 | 0.94 | 0.65 | **0.77** |
| Signs and Symptoms<br>*clinical progress* | 0.81 | 0.83 | 0.82 | 0.85 | 0.84 | **0.84** |
| Signs and Symptoms<br>*clinical examination* | 0.83 | 0.80 | 0.82 | 0.95 | 0.97 | **0.96** |
| Signs and Symptoms<br>*history of the actual condition* | 0.64 | 0.91 | 0.75 | 0.85 | 0.95 | **0.90** |
| Overall | 0.77 | 0.85 | 0.80 | 0.81 | 0.91 | **0.90** |

**Table 4** : Entity with section types as precision, recall and F1 score attributes, with naive (left) and advanced (right) body extraction. Only pairs (entities, sections) of medical interest were retained. "Vital parameters" correspond to: blood pressure, temperature, heart rate, etc. of the patient mentioned directly in the text.

## C. Detection of acute infections results (illustrative use case)

The results of the overall pipeline on document classification for acute infection causing an hospitalization are presented in Table 5. We show that the advanced body detection algorithm enabled an improvement of +7 percentage points on the F1-score for detecting acute infections.

|  | Naive body extraction | | | Advanced body extraction | | |
|---|---|---|---|---|---|---|
|  | Precision | Recall | F1-score | Precision | Recall | F1-score |
| Acute infection (n = 200 documents) | 0.80 [0.71; 0.90] | 0.74 [0.63; 0.84] | 0.77 [0.70; 0.85] | **0.83** [0.73; 0.92] | **0.85** [0.77; 0.94] | **0.84** [0.77; 0.90] |

**Table 5** : Precision, recall, and F1 score per document of the acute infection detection algorithm with and without the advanced body detection algorithm (illustrative use case). Confidence intervals were obtained using the bootstrap method.

# IV. Discussion

Health data warehouses often provide researchers with data that has undergone several stages of preprocessing from the original medical record. Ensuring the reliability of this data is critical to obtaining meaningful analyses. In this study, we introduce a lightweight algorithm for extracting clinical text from clinical PDF documents using a transformer-based model. Our approach demonstrates promising results in various aspects of information extraction, ranging from line-by-line text extraction from the body of the document to named entity recognition tasks in specific sections of the document, and to impact on a specific medical use case.

### Model architecture

Our new advanced body text extraction method preserves the original structure of clinical documents, which is of key importance to researchers and clinicians, as it allows for better interpretability, close to clinical reasoning.
Our ablation studies highlight the importance of incorporating structural features, such as contextual embeddings and relative position information encoding. Despite having a limited number of parameters, our model achieves high scores for text-body extraction. These results suggest that both contextualization through the Transformer layer and the relative position encoding in the attention mechanism play important roles in effectively capturing the structure and layout of the PDF documents, ultimately leading to more accurate segmentation and extraction.

### Impact on downstream clinical tasks

The clinical case presented in this study is essentially a proof of concept that can be generalized to many other applications. Indeed, no supervised learning was required for the detection of the clinical cases of acute infection: the extraction of medical concepts in their respective sections is a completely generic method.

Acute infection was chosen because it is a common medical complication in our autoimmune diseases, with patients often on immunosuppressive therapy. But any other phenotype could have been detected as long as it was explicitly mentioned in the text. The interest of this end-to-end method could also have been illustrated to filter out, for example, patients under certain treatments on arrival at the hospital, patients with specific family comorbidities, patients coming for a particular reason ("fever", "relapse", etc.).

## Impact on pseudonymisation

An additional advantage of the PDF segmentation method proposed in this study is its potential impact on pseudonymization. In a previous article by Tannier et al. [Tannier2023], it was shown that employing this PDF segmentation method significantly reduces the number of identifying entities in the extracted text, namely by 80%. This reduction is primarily due to the majority of identifying entities being located in side-notes, footers, and headers, which are effectively segmented and removed by our algorithm. As a result, the application of this segmentation method has been observed to increase the proportion of PDFs that have been thoroughly stripped of identifying words from 75.7% to 93.1%. This comparison was made by examining the same documents converted to text using both the naive text-extraction algorithm, i.e., a legacy rigid mask method, and the new advanced algorithm presented here. Consequently, our PDF segmentation approach not only improves the extraction of clinically relevant information but also contributes to better privacy protection through more effective pseudonymization.

## Limitations

We have not directly compared our algorithm to others like LayoutLM due to their significantly larger size and increased computational requirements. Our model, containing only 3M parameters, is fast enough to fit into the daily integration of 200,000 PDF files, the text extraction step taking approximately 15 minutes with the use of 2 GPUs and 16 cores. In contrast, the base version of LayoutLMv3, with its 133M parameters, demands considerably more computational resources and processing time. The efficiency of our algorithm in terms of time and computational resources makes it a more suitable option for large-scale PDF segmentation tasks, such as those encountered in the AP-HP's clinical data warehouse, which contains more than 120 million documents to process.

With regard to the classification of acute infection documents, we based our classification on the structure of the hospitalization reports and defined as "acute" all infectious diseases mentioned in the "clinical course" and "conclusion" section, which mainly corresponds to an actually acute infection during the hospitalization but can sometimes be simply old infections that the clinician may decide to mention.

## Perspectives

In this work, we mainly focused on body text extraction for better clinical analysis of documents; however, other applications can be found such as table and form detection, or metadata extraction from header sections for administrative and data quality validation, which will be performed in future work.

# V.  Conclusion

In this study, we developed and validated a lightweight transformer-based algorithm for extracting clinically relevant text from PDF documents. The algorithm efficiently handles various layouts of clinical PDF documents and improves the performance of downstream natural language processing tasks. Our approach demonstrates its effectiveness in preserving the original structure of clinical documents, resulting in improved interpretability and alignment with clinical reasoning, which is particularly valuable for researchers and clinicians. In addition, the computational resource efficiency of the model makes it suitable for large-scale PDF segmentation tasks in environments such as the AP-HP clinical data warehouse. The performance evaluation showed promising results in extracting medical concepts in their respective sections and on an illustrative medical use case.

# Acknowledgment

We thank the clinical data warehouse (Entrepôt de Données de Santé, EDS) of the Greater Paris University Hospitals for its support and the realization of data management and data curation tasks. We thank Xavier Tannier and Fabrice Carrat for fruitful discussions.

# Authors contribution

P.W., A.C. and B.D. had full access to all the data in the study. They take responsibility for the integrity of the data and the accuracy of the data analysis.

Concept and design: C.G., P.W., B.D., A.C., A.M., R.B.
Annotation and interpretation of data: C.G., B.D., A.C., P.W.
Manuscript drafting: C.G., P.W., R.B.
Algorithm development: P.W., B.D., A.C.
Manuscript critical proofreading C.G., P.W., A.M., B.D., R.B
Supervision: R.B.

# Conflict of Interest Disclosures

None reported.

# Data sharing

Access to the clinical data warehouse's raw data can be granted following the process described on its website: eds.aphp.fr. A prior validation of the access by the local institutional review board is required. In the case of non-APHP researchers, the signature of a collaboration contract is moreover mandatory.

# Funding/Support

This study has been supported by grants from the AP-HP Foundation.

# Role of the Funder/Sponsor

The funder was involved neither during the design and conduct of the study nor during the preparation, submission or review of the manuscript.

# Bibliography

[Liu2019] Liu X, Gao F, Zhang Q, Zhao H. Graph convolution for multimodal information extraction from visually rich documents. arXiv preprint arXiv:1903.11279. 2019 Mar 27.

[Xu2020] Xu Y, Li M, Cui L, Huang S, Wei F, Zhou M. Layoutlm: Pre-training of text and layout for document image understanding. InProceedings of the 26th ACM SIGKDD International Conference on Knowledge Discovery & Data Mining 2020 Aug 23 (pp. 1192-1200).

[Xu2022] Xu Y, Xu Y, Lv T, Cui L, Wei F, Wang G, Lu Y, Florencio D, Zhang C, Che W, Zhang M. Layoutlmv2: Multi-modal pre-training for visually-rich document understanding. arXiv preprint arXiv:2012.14740. 2020 Dec 29.

[Huang2022] Huang Y, Lv T, Cui L, Lu Y, Wei F. Layoutlmv3: Pre-training for document ai with unified text and image masking. InProceedings of the 30th ACM International Conference on Multimedia 2022 Oct 10 (pp. 4083-4091).

[Majumder2020] Majumder BP, Potti N, Tata S, Wendt JB, Zhao Q, Najork M. Representation learning for information extraction from form-like documents. Inproceedings of the 58th annual meeting of the Association for Computational Linguistics 2020 Jul (pp. 6495-6504).

[Kim2022] Kim G, Hong T, Yim M, Nam J, Park J, Yim J, Hwang W, Yun S, Han D, Park S. Ocr-free document understanding transformer. InComputer Vision–ECCV 2022: 17th European Conference, Tel Aviv, Israel, October 23–27, 2022, Proceedings, Part XXVIII 2022 Oct 20 (pp. 498-517). Cham: Springer Nature Switzerland.

[Yu2021] Yu W, Lu N, Qi X, Gong P, Xiao R. PICK: processing key information extraction from documents using improved graph learning-convolutional networks. In2020 25th International Conference on Pattern Recognition (ICPR) 2021 Jan 10 (pp. 4363-4370). IEEE.

[Pomares2019] Pomares-Quimbaya A, Kreuzthaler M, Schulz S. Current approaches to identify sections within clinical narratives from electronic health records: a systematic review. BMC medical research methodology. 2019 Dec;19(1):1-20.

[Kropf2017] Kropf S, Krücken P, Mueller W, Denecke K. Structuring legacy pathology reports by openEHR archetypes to enable semantic querying. *Methods Inf Med*. 2017;56(3):230–7.

[Zweigenbaum2013] Zweigenbaum P, Deléger L, Lavergne T, Névéol A, Bodnari A. A supervised abbreviation resolution system for medical text. *In: Working Notes for CLEF Conference, Valencia, Spain*, September 23-26, volume 1179 of CEUR Workshop Proceedings; 2013

[Edinger2017] Edinger T, Demner-Fushman D, Cohen AM, Bedrick S, Hersh W. Evaluation of clinical text segmentation to facilitate cohort retrieval. *AMIA Annu Symp*. 2017;2017:660–9.

[PyMuPDF] Liu, R., & McKie, J. X.., PyMuPDF. Available at: http://pymupdf.readthedocs.io/en/latest/ [Accessed april 25, 2023]

[Vaswani2017] Vaswani, A., Shazeer, N., Parmar, N., Uszkoreit, J., Jones, L., Gomez, A. N., Kaiser, L., & Polosukhin, I. (2017). Attention is all you need. *Advances in Neural Information Processing Systems*, *2017-Decem*, 5999–6009.

[He2020] He, P., Liu, X., Gao, J., & Chen, W. (2020). *DeBERTa: Decoding-enhanced BERT with Disentangled Attention*. http://arxiv.org/abs/2006.03654

[Wajsbürt2021] Perceval Wajsbürt. Extraction and normalization of simple and structured entities in medical documents. Santé publique et épidémiologie. Sorbonne Université, 2021. English. ⟨NNT : 2021SORUS541⟩. ⟨tel-03624928v2⟩

[Gérardin2022] Gérardin C, Wajsbürt P, Vaillant P, Bellamine A, Carrat F, Tannier X. Multilabel classification of medical concepts for patient clinical profile identification. Artificial Intelligence in Medicine. 2022 Jun 1;128:102311.

[Yu2020] Yu, Juntao, Bernd Bohnet, and Massimo Poesio. "Named entity recognition as dependency parsing." *arXiv preprint arXiv:2005.07150* (2020).

[Bojanowski2017] Bojanowski, Piotr, et al. "Enriching word vectors with subword information." *Transactions of the association for computational linguistics* 5 (2017): 135-146.

[Devlin2018] Devlin, Jacob, et al. "Bert: Pre-training of deep bidirectional transformers for language understanding." *arXiv preprint arXiv:1810.04805* (2018).

[MeSH]  https://www.ncbi.nlm.nih.gov/mesh/

[Weiskopf 2013] Weiskopf NG, Hripcsak G, Swaminathan S, Weng C. Defining and measuring completeness of electronic health records for secondary use. J Biomed Inform. 2013;46(5):.

[Tannier2023] Tannier, X., Wajsbürt, P., Calliger, A., Dura, B., Mouchet, A., Hilka, M., & Bey, R. (2023). *Development and validation of a natural language processing algorithm to pseudonymize documents in the context of a clinical data warehouse*. *arXiv preprint arXiv:2303.13451 (2023)*

[PyTorch] Paszke A, Gross S, Massa F, Lerer A, Bradbury J, Chanan G, et al. PyTorch: An Imperative Style, High-Performance Deep Learning Library. In: Advances in Neural Information Processing Systems 32 [Internet]. Curran Associates, Inc.; 2019. p. 8024–35.

[MyCray2001] McCray, A. T.; Burgun, A. & Bodenreider, O.Aggregating UMLS semantic types for reducing conceptual complexity.Studies in health technology and informatics, 2001, 84, 216-220

# Supplementary materials

Supplementary Table 1: Lines repartition for the Train and Test datasets.

| Line label | Train | Test |
|---|---:|---:|
| **Body** | 11,841 | 3,275 |
| **Footer** | 602 | 136 |
| **Header** | 3,748 | 927 |
| **Left note** | 4,903 | 1,325 |
| **Page index** | 232 | 67 |
| **others** | 608 | 185 |
| **Signature** | 245 | 74 |
| **Title** | 267 | 59 |
| **Total documents** | 215 | 57 |
| **Total pages** | 423 | 115 |
| **Total lines** | 22,446 | 6,048 |

Supplementary Table 2: Details of the section's name (In French) for the rule-based section detection.

| Section type | Term |
|---|---|
| Conclusion | conclusion |
| Conclusion | Au total |
| Conclusion | conclusion médicale |
| Conclusion | Synthèse du séjour |
| Conclusion | Synthèse |
| Conclusion | conclusion de sortie |
| Conclusion | syntese medicale / conclusion |
| Conclusion | synthese medicale |
| Conclusion | conclusion consultation |
| Conclusion | diagnostic retenu |
| | |
| Histoire de la maladie | histoire de la maladie |
| Histoire de la maladie | histoire recente |
| Histoire de la maladie | histoire recente de la maladie |
| Histoire de la maladie | rappel clinique |
| Histoire de la maladie | resume |
| Histoire de la maladie | resume clinique |
| Histoire de la maladie | histoire de la maladie - explorations |
| Histoire de la maladie | histoire de la maladie actuelle |
| Histoire de la maladie | évènements récents |
| Histoire de la maladie | evolution depuis la dernière consultation |
| Histoire de la maladie | Résumé clinique - Histoire de la maladie |
| Histoire de la maladie | histoire clinique |
| Histoire de la maladie | rappel |
| Histoire de la maladie | Rappel de la conclusion de la précédente consultation |
| Histoire de la maladie | Evaluation des effets secondaires en intercure |
| Histoire de la maladie | historique |
| Histoire de la maladie | Pour mémoire |
| Histoire de la maladie | Anamnèse |
| Histoire de la maladie | Résumé des séances |
| | |
| Indication | indication |
| Indication | Contexte clinique |
| Indication | renseignements cliniques |
| Indication | Motif de l'examen |
| | |
| Résultats | résultats |
| | |
| Motif | motif |
| Motif | motif médical |
| Motif | motif d'hospitalisation |
| Motif | motif de la consultation |
| Motif | motif de consultation |
| Motif | motif de la consultation recueilli par l'iao |
| Motif | motif de l'hospitalisation |
| Motif | Motif de prise en charge |
| Motif | Motif d'admission |
| Motif | Motif de présentation |
| | |
| Examens complémentaires | examens complémentaires |
| Examens complémentaires | examens complémentaires (résultats et commentaires) |
| Examens complémentaires | biologie |
| Examens complémentaires | ECG |
| Examens complémentaires | biochimie |
| Examens complémentaires | Scanner |
| Examens complémentaires | Radiographie |
| Examens complémentaires | résultats de biologie |
| Examens complémentaires | imagerie |
| Examens complémentaires | PET-scanner |
| Examens complémentaires | examen complémentaire réalisé à l'entrée |
| Examens complémentaires | echocardiographie |
| Examens complémentaires | interpretation des examens complementaires |
| Examens complémentaires | examen(s) complementaire(s) |
| Examens complémentaires | examens complementaires a l'entree |
| Examens complémentaires | examens complementaires realises pendant le sejour |
| Examens complémentaires | examens para-cliniques |
| Examens complémentaires | biologie à l'entrée |
| Examens complémentaires | examen histologique |
| Examens complémentaires | examens à l'entrée |
| Examens complémentaires | anatomo-pathologie |
| Examens complémentaires | Résultat du bilan |
| Examens complémentaires | Diagnostic histopathologique |
| Examens complémentaires | Examen cytologique |
| Examens complémentaires | Test au Synacthène |
| Examens complémentaires | Mammographie |
| Examens complémentaires | Etude immunohistochimique |
| Examens complémentaires | RADIOGRAPHIE THORACIQUE |
| Examens complémentaires | Examens d'imagerie |
| Examens complémentaires | Bilan génétique |
| Examens complémentaires | EXAMENS PARACLINIQUES |
| Examens complémentaires | Ionogramme |
| | |
| Autres | avis specialises |
| Autres | document remis au patient |
| Autres | dossier social |
| Autres | décision d'orientation |
| Autres | risque infectieux |
| Autres | liste des prescriptions demandées |
| Autres | action iao |
| Autres | actes ide réalisés à l'accueil |
| Autres | devenir réel du patient |
| Autres | dossier traumatologie |
| Autres | affection exonerante |
| Autres | modalites de sortie |
| Autres | documents remis au patient |
| Autres | rendez-vous pris |
| Autres | mode d'arrivée |
| Autres | planification des soins / suites à donner |
| Autres | CONSULTATION POST URGENCE |
| Autres | relecture du dossier |
| Autres | décision de la rcp |
| Autres | suivi post HdJ |
| Autres | codage |
| Autres | planification des soins |
| Autres | avis psychiatrique |
| Autres | Documents de sortie |
| Autres | Rendez-vous programmés |
| Autres | SYNTHESE INFIRMIERE |
| Autres | SYNTHESE INFIRMIERE A LA SORTIE |
| Autres | TRANSMISSION IDE |
| Autres | COMMENTAIRES |
| Autres | SCORES |
| Autres | Soins de rééducation |
| Autres | Soins infirmiers |
| | |
| Examen clinique | examen clinique initial |
| Examen clinique | examen clinique |
| Examen clinique | constantes initiales |
| Examen clinique | examen clinique à l'entrée |
| Examen clinique | examen dermatologique |
| Examen clinique | Evaluation de l'état général |
| Examen clinique | EXAMEN D'ENTRÉE |
| Examen clinique | Contexte clinique du patient |
| Examen clinique | CLINIQUE |
| Examen clinique | EXAMEN CLINIQUE A L'ADMISSION |
| | |
| Traitement | attitude therapeutique initiale |
| Traitement entrée | traitement à l'entrée |
| Traitement entrée | traitement en cours |
| Traitement entrée | traitement habituel |
| Traitement entrée | traitement actuel |
| Traitement de sortie | Traitement de sortie |
| Traitement de sortie | traitement et ordonnance de sortie |
| Traitement de sortie | ordonnance de sortie |
| Traitement | traitement |
| Traitement | prescriptions |
| Traitement | autres prescriptions |
| Traitement | prescriptions relatives au traitement de l'affection de longue durée reconnue |
| Traitement de sortie | prescriptions de sortie |
| Traitement de sortie | prescriptions medicales de sortie |
| Traitement | prescriptions sans rapport avec l'affection de longue durée |
| Traitement | TRAITEMENTS PRESCRITS |
| Traitement | traitements en cours ou d'administration recente |
| Traitement | médicaments |
| Traitement de sortie | prescriptions à l'issue de la consultation |
| Traitement | Traitement et surveillance |
| Traitement | Anti-infectieux |
| Traitement | Traitement spécifique |
| | |
| Antécédents | antécédents |
| Antécédents | atcd |
| Antécédents | antécédents médicaux |
| Antécédents | antécédents - allergies |
| Antécédents | antecedents medicaux et chirurgicaux |
| Antécédents | antecedents personnels |
| Antécédents familiaux | antécédents familiaux |
| Antécédents | antécédents chirurgicaux |
| Allergies | allergie |
| Allergies | Allergies connues |
| | |
| Antécédents | facteurs de risque cardiovasculaires |
| Antécédents | autres antecedents medicaux ou chirurgicaux |
| Antécédents | antécédents gynéco-obstétriques |
| Antécédents | sur le plan des facteurs de risque cardiovasculaires |
| Antécédents | Vaccinations |
| Antécédents | antécédents cardiaques |
| Antécédents | diagnostics associés |
| Antécédents | facteurs de risques |
| Antécédents | médicaux |
| Antécédents | chirurgicaux |
| Antécédents | Antécédents et mode de vie |
| Antécédents | Antécédents psychiatriques |
| Antécédents | Antécédent médico-chirurgical |
| Antécédents | vaccins |
| | |
| Evolution | Evolution clinique |
| Evolution | evolution |
| Evolution | évènements recensés au cours de ce séjour |
| Evolution | Evolution dans le service |
| Evolution | Conclusion à l'entrée |
| Evolution | Prise en charge |
| Evolution | Traitement / Evolution dans le service |
| Evolution | Evolution post greffe : |
| Evolution | HYPOTHESES DIAGNOSTIQUES : |
| Evolution | Suites opératoires : |
| | |
| Constantes | Paramètres vitaux à l'entrée |
| Constantes | parametres vitaux initiaux |
| Constantes | constantes initiales |
| Constantes | constantes |
| Constantes | dernières constantes |
| Constantes | donnees biometriques et parametres vitaux a l'entree |
| Constantes | parametres vitaux et donnees biometriques a l'entree |
| Constantes | ACCUEIL IAO |
| Constantes | Paramètres vitaux à l'accueil des urgences |
| Constantes | PARAMETRES DE SURVEILLANCE |
| | |
| Mode de vie | mode de vie |
| Mode de vie | habitus |
| Mode de vie | mode de vie - scolarite |
| Mode de vie | situation sociale, mode de vie |
| Mode de vie | contexte familial et social |
| Mode de vie | Environnement familial |
| Mode de vie | MDV |
| Mode de vie | Mode de vie et éléments biographiques |